\def\year{2022}\relax
\documentclass[letterpaper]{article} 
\usepackage{aaai22}  
\usepackage{times}  
\usepackage{helvet}  
\usepackage{courier}  
\usepackage[hyphens]{url}  
\usepackage{graphicx} 
\usepackage{natbib}  
\usepackage{caption} 
\DeclareCaptionStyle{ruled}{labelfont=normalfont,labelsep=colon,strut=off} 
\usepackage{algorithm}
\usepackage{algorithmic}
\usepackage{amsthm}

\usepackage{bbm}
\usepackage{amsmath}
\usepackage{subcaption}
\usepackage{amssymb}
\newtheorem{theorem}{Theorem}[section]

\usepackage{booktabs}

\usepackage{newfloat}
\usepackage{listings}
\floatstyle{ruled}
\newfloat{listing}{tb}{lst}{}
\floatname{listing}{Listing}
\usepackage{comment}
\title{Counterfactual Explanations via Locally-guided Sequential Algorithmic Recourse}
\author {
    Edward A. Small,\textsuperscript{\rm 1}
    Jeffrey N. Clark,\textsuperscript{\rm 2}
    Christopher J. McWilliams,\textsuperscript{\rm 2}
    Kacper Sokol~\textsuperscript{\rm 3}\\
    Jeffrey Chan,\textsuperscript{\rm 1}
    Flora D. Salim,\textsuperscript{\rm 4}
    Raul Santos-Rodriguez~\textsuperscript{\rm 2}
}
\affiliations {
    \textsuperscript{\rm 1} Royal Melbourne Institute of Technology\\
    \textsuperscript{\rm 2} University of Bristol\\
    \textsuperscript{\rm 3} ETH Z{\"u}rich\\
    \textsuperscript{\rm 4} University of New South Wales\\
    edward.small@student.rmit.edu.au, enrsr@bristol.ac.uk
}

\usepackage{bibentry}

\begin{document}

\maketitle

\begin{abstract}
Counterfactuals operationalised through algorithmic recourse have become a powerful tool to make artificial intelligence systems explainable. Conceptually, given an individual classified as $y$ -- the factual -- we seek actions such that their prediction becomes the desired class $y^\prime$ -- the counterfactual. This process offers algorithmic recourse that is (1) easy to customise and interpret, and (2) directly aligned with the goals of each individual. However, the properties of a ``good'' counterfactual are still largely debated; it remains an open challenge to effectively locate a counterfactual along with its corresponding recourse. Some strategies use gradient-driven methods, but these offer no guarantees on the feasibility of the recourse and are open to adversarial attacks on carefully created manifolds. This can lead to unfairness and lack of robustness. Other methods are data-driven, which mostly addresses the feasibility problem at the expense of privacy, security and secrecy as they require access to the entire training data set. Here, we introduce \textsc{LocalFACE}, a model-agnostic technique that composes feasible and actionable counterfactual explanations using locally-acquired information at each step of the algorithmic recourse. Our explainer preserves the privacy of users by only leveraging data that it specifically requires to construct actionable algorithmic recourse, and protects the model by offering transparency solely in the regions deemed necessary for the intervention.
\end{abstract}

\section{Introduction}

The adoption of complex artificial intelligence (AI) and machine learning (ML) methods has soared in recent years. These methods often constitute automated systems that are difficult to fully understand or interpret, earning them the \emph{black box} moniker. Consequently, it can be difficult to understand how a black-box model is exploiting data to make decisions, where bias and unfairness can creep into the system, and how robust and safe a model is to use. 

Despite these unknowns, black-box models are used in production in many high stakes domains such as healthcare and finance~\cite{rudin2019stop}. Thus, there is a desperate need for both inherently interpretable models and explainability techniques that both contribute to explainable AI (XAI). Interpretable modelling is often thought of in the design phase of creating an automated system, with the goal of having transparency as a part of the objective of the system. A prototypical example of this is a glass-box predictive model that offers full transparency but can often come at the cost of requiring a high level of expertise due to the complexity of the model itself~\cite{sokol2023reasonable}. Explainable methods, on the other hand, are used \emph{post factum} -- we are given a black box and asked to extract some rationale behind its decision-making process.

\begin{figure*}[t]
    \centering
    \includegraphics[width=1\textwidth]{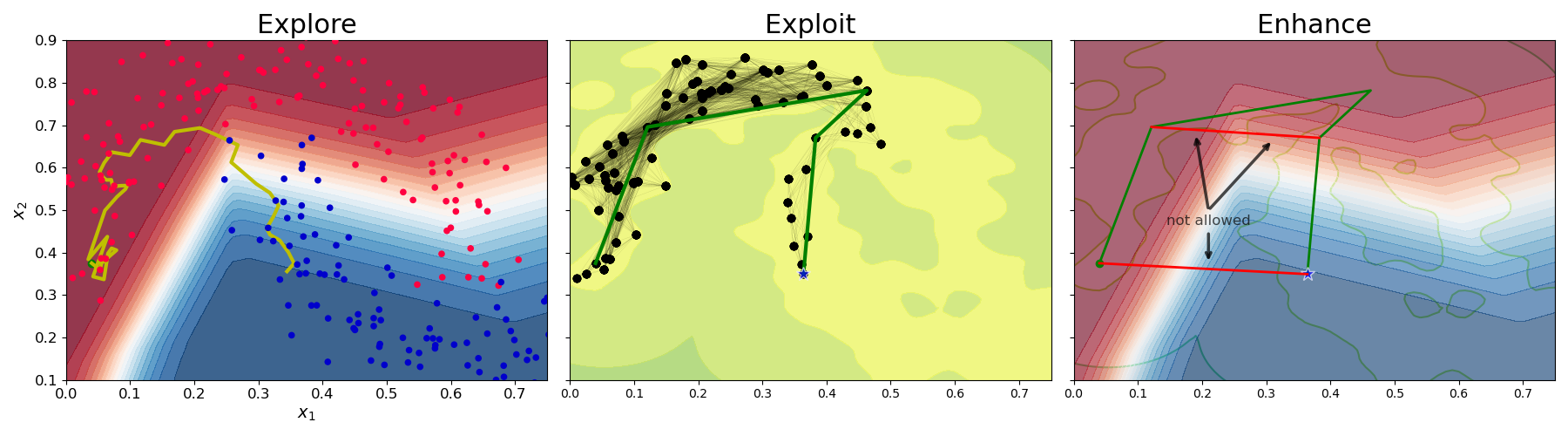}
    \caption{A break down of the \textsc{LocalFACE} pipeline on a 2-D data set (two moons). The \textit{explore} stage finds an accessible counterfactual by searching through data guided by increasing probabilities from the decision manifold (red to blue); the \textit{exploit} stage generates a graph of minimal size by searching for relevant data points (which become vertices) along the data-dense route (yellow) from the factual to the counterfactual point; the \textit{enhance} stage finds optimal recourse through the graph such that the path never leaves the viable region (bounded area).}
    \label{fig:abstract}
\end{figure*}

A particularly powerful suite of techniques in XAI is counterfactual explainability enhanced with algorithmic recourse~\cite{10.1145/3442188.3445899}
. A counterfactual is any outcome that could have occurred but did not; algorithmic recourse captures the actions one needs to take to make the counterfactual occur. Together they form a counterfactual explanation (CFE), often encapsulated by a statement such as ``\textit{if I had done $Z$ actions, $Y$ would have happened instead of $X$}''. This makes them attractive for two distinct reasons: first, from the user's perspective, they offer explicit instructions on how to improve one's outcome -- very little interpretation or expertise is required; second, from an engineering perspective, the model can remain private and opaque -- only very specific and relevant information is revealed to the user.
Not only can CFEs be used for inspecting a model during its training and validation~\cite{teney2020learning}, but they can also be used to expose bias and discrimination in a model by assessing the algorithmic recourse offered to different protected groups~\cite{slack2021counterfactual}.

When developing CFEs we need to consider their two aspects: finding the counterfactual and finding the trajectory from the factual to the counterfactual. Some methods, such as counterfactual optimisation~\cite{WachterSandra2017CEWO} and DICE~\cite{Mothilal}, use the model directly in order to extract a counterfactual, whereas others, such as FACE~\cite{10.1145/3375627.3375850}, use data to find the necessary intervention. However, these methods have some limitations: (1) gradient-based techniques find close counterfactuals, but offer no guarantees on the feasibility of the recourse and are susceptible to model changes and adversarial attacks; and (2) data-driven methods cannot directly find counterfactuals, and must access the entire data set, which raises privacy concerns.

To address these shortcomings we introduce \textsc{LocalFACE}, a novel method that relies on a three-step process, depicted in Figure~\ref{fig:abstract}, to: (1) find a counterfactual (explore), (2) extract the needed local information (exploit), and (3) optimise the feasible path (enhance)\footnote{\url{https://github.com/Teddyzander/localFACE}}. Our approach relies on moving through areas of dense data, solving the feasibility problem, but does not require prior access to the entire data set, preserving privacy. It is model-agnostic and can retrieve diverse counterfactuals together with the associated interventions by tuning density and distance parameters and/or enforcing sparsity on the feature vector. The path from factual to counterfactual does not rely directly on the model itself and is therefore less susceptible to changes in the model. We show this on a \textit{readiness for discharge} (RFD) case study where we generate feasible counterfactual paths for patients labelled as not ready for discharge (NRFD).

\section{Preliminaries}

\subsection{Notation}

We take $\mathcal{X}\subseteq\mathbb{R}^d$ to be a $d$-dimensional ($d \in \mathbb{N}$) input space with $\mathcal{Y}$ being the corresponding label space, which without loss of generality we assume to be binary, i.e., $\mathcal{Y} = \{0, 1\}$. $X = \{x_i\}_{i=1}^n$ represents $n \in \mathbb{N}$ samples $x_i \in \mathcal{X}$ from the input space according to a density $p$.
The scalar decision function $f:\mathcal{X}\mapsto\mathbb{R}$ is defined on the input space, as is the probability density function $f_p:\mathcal{X}\mapsto[0,1]$, and we say there exists some classification process $g:\mathbb{R}\mapsto\mathcal{Y}$, e.g., based on thresholding. $G=(V, E, W)$ is an undirected graph with $V$ vertices, $E$ edges and $W$ weights, where $e_{i,j}=e_{j,i}$ is an edge that connects vertices $v_i$ and $v_j$ with a weight $w_{i,j}=w_{j,i}$. We use $T_f$ and $T_p$ to describe threshold criteria on the decision function $f$ and the density function $p$ respectively.
Additionally, we define $x$ as the factual (starting) point, $\hat{y}$ as the factual outcome, $x^\prime$ as the counterfactual point and $\hat{y}^\prime$ as the counterfactual outcome. $Z\in\mathbb{R}^{d\times k}$, called the \emph{recourse matrix}, contains the list of $k$ steps needed to transform $x$ into $x^\prime$.

\subsection{Counterfactual Explanations}

The aim of a CFE is to discover what actions $Z$ are needed to move from the factual $x$ over the decision boundary $T_f$, so if
\begin{equation}
    x^\prime = x + \sum_{i=1}^{k} z_i \quad \text{where} \quad Z=[z_1,\ldots,z_k]
    \label{eq:steps}
\end{equation}
then
\begin{equation*}
    \begin{aligned}
        f(x) < T_f &\implies g\big(f(x)\big) = \hat{y} \\ 
         f(x^\prime) \geq T_f &\implies g\big(f(x^\prime)\big) = \hat{y}^\prime
    \end{aligned}
    \text{~.}
\end{equation*}
CFEs can be applied to a multitude of problems and have the added advantage that the underlying model does not need to be interpretable in order to get meaningful explanations. They are also highly flexible, as they can weakly capture complex relationships and dynamics that can occur between features in $\mathcal{X}$ that would be present in a causal model, which itself can be prohibitively difficult to construct. As such, most CFE methods look to find desirable candidates for $x^\prime$ with respect to the model itself.

\section{Sequential Algorithmic Recourse}

Most methods consider the transition from factual to counterfactual to be a single step, i.e., $z=x^\prime - x$. However, this representation of algorithmic recourse has numerous shortcomings. Unless $z$ is sparse, presenting it as a single step may be overwhelming and seemingly infeasible due to its sheer complexity, given that it implies that most features need altering. We could therefore look to break up $z$ into a series of $k$ steps $Z$ such that Equation~\ref{eq:steps} holds, enforcing sparsity on the components of $Z$ as a proxy for step complexity, i.e., few feature changes per step~\cite{sokol2023navigating}. However, there are an infinite number of ways to achieve this. Furthermore, because many CFE methods ignore feature covariance there is no guarantee that every (or any) step in the chosen realisation of $Z$ is actually feasible, and there is no way to know what order the steps should be completed (or whether the order matters at all). For example, a readiness for discharge model may suggest that a patient requires higher oxygen saturation levels and to come off supplementary oxygen in order to be discharged. Performing one of these actions before the other may be simpler or give higher odds of success.

The FACE method~\cite{10.1145/3375627.3375850} was developed to solve these issues. FACE uses data density as a proxy for how feasible an action is and looks to create a path through data-dense regions where each step is no bigger than a threshold $\epsilon \in \mathbb{R}^+$. As such, the final recourse matrix $Z$ is designed in such a way that each column $\lVert z_i \rVert \leq \epsilon$ and the path moves through the densest regions of data by following pre-existing instances. Nonetheless, FACE fails to address certain desirable aspects of algorithmic recourse:
\begin{enumerate}
    \item by using the distance threshold $\epsilon$ to enforce small steps, any individual that exists in a sparse data region is overlooked by the algorithm, i.e., it is not linked to the graph;
    \item since regions of interest are not known in advance, FACE requires access to data that span the entire manifold, thus raising privacy concerns; and
    \item FACE is not designed to find feasible counterfactuals -- it simply uncovers a feasible path to a given counterfactual within the given data set.
\end{enumerate}
To address these shortcomings, we propose a method that finds a nearby feasible counterfactual, creates a local graph of the data between the factual and the counterfactual, and finds optimal recourse through regions of consistent density. Separating the construction of algorithmic recourse into these three distinct parts allows our method to only require access to data deemed \emph{relevant}. Therefore, only the areas that the recourse passes through become transparent.

\begin{figure}[t]
    \centering
    \includegraphics[width=0.4\textwidth]{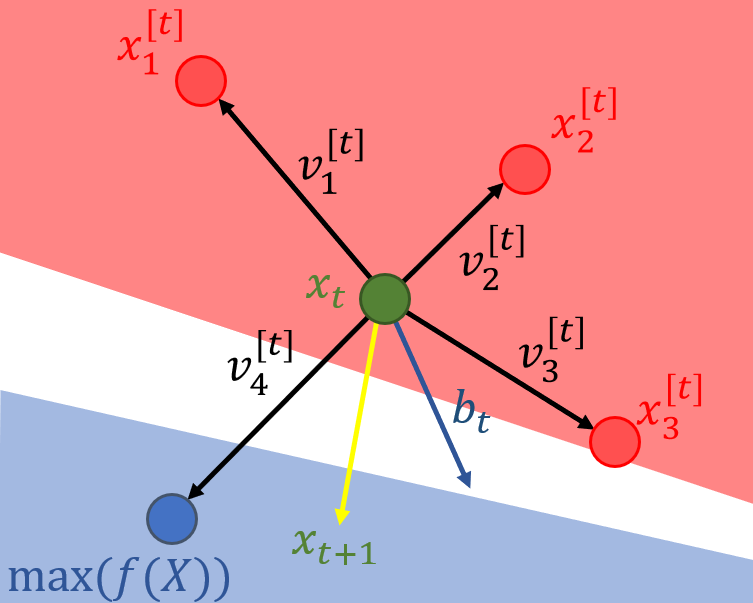}
    \caption{An example for finding $x_{t+1}$ for 4 nearest neighbours using momentum $b_t$.If $f(x_{t}) \geq T_f$ then $x^\prime = x_t$.}
    \label{fig:knn}
\end{figure}

\subsection{Step 1: Explore}

Whilst suitable methods to locate counterfactual candidates $x^\prime$ are abundant, their objectives are misaligned with our CFE desiderata. Therefore, we propose an alternative search technique that maximises the likelihood that a path between $x$ and $x^\prime$ exists. 

At any iteration of our algorithm -- denoted by a point $x_t$ -- we seek a candidate point that is likely to be in the direction of the chosen counterfactual class by finding the $k$ nearest neighbours $x_i \in G_t \subset X$ of $x_t$ and considering
\begin{equation}
    x_{t+1} = x_t + \frac{\max_i(\frac{f(x_i)}{1 + \lVert u_i \rVert}) - x_t + b_t}{2} \quad \forall x_i\in G_t
    \label{eq:xt}
\end{equation}
where $u_i = x_i - x_t$ and $b_t$ is the average direction of the last $m$ steps, which can be understood here as the \emph{momentum}:
\begin{equation}
    b_t = \frac{1}{m}\sum_{i=(t-m)}^t z_i
    \text{~.}
    \label{eq:mom}
\end{equation}

\begin{algorithm}[t!]
\caption{Find viable counterfactual $x^\prime$. \label{alg:momentum}}
\begin{algorithmic}[1]
    \REQUIRE
        starting point $x_0$; decision function $f$; KD-Tree of data set $T_X$; decision threshold $T_f$; $m$ history for momentum; $k$ for nearest neighbours.
    \ENSURE
        best counterfactual candidate $x^\prime$. 
    \STATE $t\gets 0$
    \STATE $x_t \gets x_0$ 
    \WHILE{$f(x_t) < T_f$}
        \STATE $G_t\gets T_X(x_t)$ \COMMENT{get $k$-NN by querying $T_X$}
        \IF{$m > t > 0$}
            \STATE $b_t \gets \frac{1}{t} \sum_{i=1}^t (Z_{i} - Z_{i-1})$
        \ELSE
            \STATE $b_t \gets$ Equation~\ref{eq:mom}
        \ENDIF
        \STATE $x_{t+1} \gets$ Equation~\ref{eq:xt}
        \STATE $Z_{t+1} \gets x_{t+1} - x_t$
        \STATE $T_X \gets T_X - x_{t+1}$ \COMMENT{update tree}
        \STATE $t \gets t + 1$
    \ENDWHILE
    \STATE $x^\prime \gets x_t$
\end{algorithmic}
\end{algorithm}

\begin{figure}[t]
    \centering
    \includegraphics[width=0.45\textwidth]{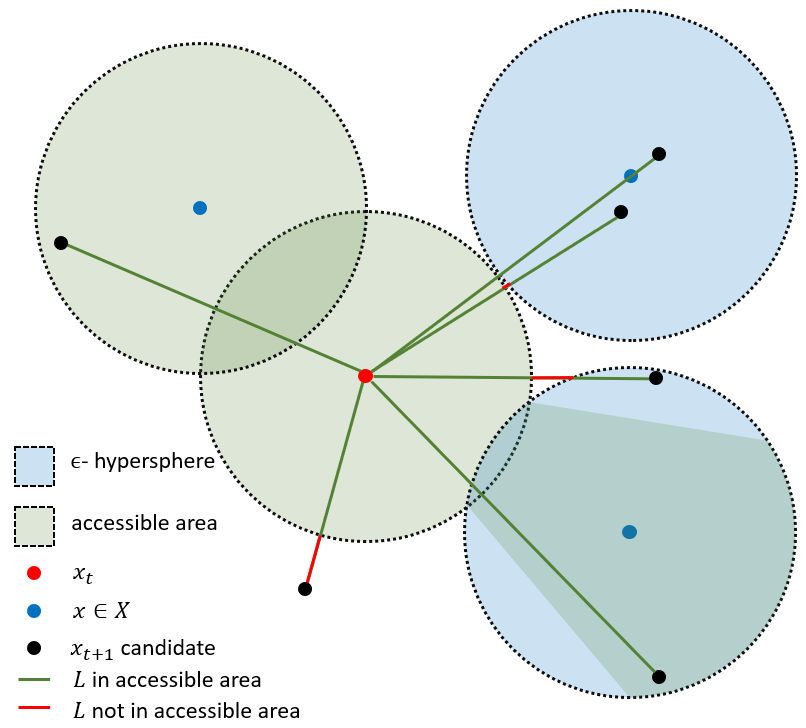}
    \caption{When finding the next step in counterfactual search we do not require that $x_{t+1} \in X$, but the line $L_{t,t+1}$ that connects $x_t$ and $x_{t+1}$ must adhere to $\max(\lvert L_{t,t+1} - X \rvert) \leq \epsilon$ (the green area) -- see Theorem~\ref{the:pythagorean}.}
    \label{fig:dist}
\end{figure}

A single step of this process is shown in Figure~\ref{fig:knn} and captured in Algorithm~\ref{alg:momentum}. We repeat this process until $f(x_t)>T_f$, in which case $x^\prime = x_t$. During the search, the best next step is allowed to slightly deviate away from the given data, but only as so far that the maximum distance any point on the line connecting two vertices is less than $\epsilon$ from any known data point in $X$, as shown in Figure~\ref{fig:dist}. Such a constraint allows for some flexibility in the search, but ensures that $x^\prime$ is accessible from $x$. This is implemented through operationalisation of Theorem~\ref{the:pythagorean} (proof in Appendix~\ref{sec:min_angle}), which captures the accessible local space the next step must exist in from any given point. If $x^\prime$ is already known, this stage can be skipped.

\begin{theorem}[Maximum Angle of Deviation]
\label{the:pythagorean}
Consider two hyper-spheres of radius $\epsilon$ centred respectively on the points $x_1,x_2\in\mathcal{X}$ such that if $d:\mathcal{X}\times\mathcal{X}\mapsto\mathbb{R}$ is a distance function, then $d(x_1,x_2) \leq 2\epsilon$. Consider $L_x$ to be the line connecting $x_1$ and $x_2$, and $L$ to be any line that connects $x_1$ to $x_t$ such that $d(x_2,x_t) \leq \epsilon$. If $\arccos{(\phi)}$ is the angle at $x_1$ between $L$ and $L_x$, then in order for $d(x, x_1) \leq \epsilon$ or $d(x, x_2) \leq \epsilon$ for all $x$ on $L$ then the following must hold
\begin{equation*}
    \phi \geq 
    \frac{1}{2}\Big(1 + \frac{d}{\epsilon}\Big)
    \text{~,}
\end{equation*}
where
\begin{equation*}
    \phi = \frac{(x_2 - x_1) \cdot (x_t - x_1)}{\lVert x_2 - x_1 \rVert \lVert x_t - x_1 \rVert}
    \text{~.}
\end{equation*}
\end{theorem}

\subsection{Step 2: Exploit}

Given a known counterfactual $x^\prime$, we now look to compose a graph $G$ of points $V \subseteq X$ connecting $x$ to $x^\prime$. To this end, we exploit the knowledge that we always know the direction of the counterfactual at any step $x_t$ of the process, which is simply computed as $x^\prime - x_t$, therefore
\begin{equation}
\begin{aligned}
    v_{t+1} = x_{i} \qquad \text{where} \\
    \max_{x_i}
    \bigg (\frac{1+\frac{(x_{i} - v_t)\cdot(x^\prime - v_t)}{\lVert x_{i} - v_t \rVert \lVert x^\prime - v_t \rVert}}{2} \cdot \frac{1}{\lvert L_{i,t}\rvert}\int_{L_{i,t}}f_p(r)dr \bigg )
    \label{eq:exploit}
\end{aligned}
\end{equation}
for all $x_i\in G_t$. In other words, each node $v_{t+1}$ we add to the graph $G$ ought to be aligned with the direction from the previous node and the counterfactual (term 1), weighted by the average density along the path (term 2). We could consider allowing nodes based on instances we have not observed earlier $V_i\not\subset X$ provided that the conditions specified in Figure~\ref{fig:dist} are satisfied. To estimate the average density, we can simply take $q \in \mathbb{N}$ samples along the line. As $q\to\infty$,
\begin{equation*}
    \frac{1}{\lvert L_{i,j}\rvert}\int_{L_{i,j}}f_p(r)dr \approx \frac{\sum_{i=0}^q f_p(\frac{q-i+1}{q+1}V_i + \frac{i}{q+1} V_{j})}{q+1}
    \text{~.}
\end{equation*}
We call this value $\mathcal{D}_{i,j}$ and it captures the average probability along a line.

\begin{algorithm}[t]
\caption{Generate local graph.\label{alg:exploit}}
\begin{algorithmic}[1]
    \REQUIRE
        starting point $x_0$; counterfactual $x^\prime$; maximum distance $\epsilon$ or $k$-NN; KD-Tree of data set $T_X$; $f_p$ probability density function.
    \ENSURE
        Local graph $G$.
    \STATE $t\gets 0$
    \STATE $x_t \gets x_0$
    \STATE $Z_t \gets x_t$ 
    \STATE $V_t \gets x_t$
    \WHILE{$x_t \neq x^\prime$}
        \STATE $G_t \gets T_X(x_t, \epsilon; k)$ \COMMENT{get all points near $x_t$}
        \STATE $x_{t+1} \gets$ Equation~\ref{eq:exploit} 
        \STATE $T_X \gets T_X - x_{t+1}$ \COMMENT{update tree}
        \STATE $v_{t+1} \gets x_{t+1}$
        \STATE $t \gets t + 1$
        \FOR{$i=0$; $i>t$; $i=i+1$}
            \IF{$f_p(x) > T_p$}
                \STATE $w_{i,t} \gets$ \COMMENT{Equation~\ref{eq:weights}}
            \ELSE
                \STATE $w_{i,t} \gets 0$
            \ENDIF
        \ENDFOR
    \ENDWHILE
\end{algorithmic}
\end{algorithm}

After a new node $v_{t+1}$ is added to the graph, we must consider which other nodes currently in $G$ it may have a viable edge to. Here, FACE limits the distance $\epsilon$ for adding edges in the global graph. However, there are three fundamental shortcomings of such a strategy:
\begin{enumerate}
    \item small $\epsilon$ yields an improved estimate of the geodesic but excludes individuals who exist in sparse data regions i.e., any underrepresented population;
    \item small $\epsilon$ creates a recourse matrix $Z$ whose number of steps $k$ is exceedingly large; and
    \item in principle we can construct large steps that meet the feasibility criteria by considering the density directly as demonstrated by Figure~\ref{fig:dist}.
\end{enumerate}
Therefore, the distance between nodes is fundamentally irrelevant provided that the space between them is deemed to be traversable according to some criteria. For \textsc{LocalFACE} we consider two possible weight calculations:
\begin{equation}
    \begin{aligned}
        w_{i,j} &=  
        \begin{cases}
            \lVert v_i - v_j \rVert \qquad &\text{if} \quad f_p(x) > T_p, \;  \forall x \in L_{i,j} \\
            0 \quad &\text{otherwise}
        \end{cases}
        \\
        w_{i,j} &=
        \begin{cases}
            \mathcal{D}_{i,j}\lVert v_i - v_j \rVert \qquad &\text{if} \quad \mathcal{D}_{i,j} > T_p \\
            0 \quad &\text{otherwise}
            \label{eq:weights}
        \end{cases}
    \end{aligned}
    \text{~.}
\end{equation}
The first weight calculation is \textit{strict} as it requires all points on the line connecting two vertices to always have a density greater than $f_p$. Because of this, we do not use the density in the weight itself, as the existence of $e_{i,j}$ implies it meets the density criteria. The second weight calculation is an \text{average} as we simply require the average density along the line connecting two vertices to be greater than the threshold. This is significantly more flexible than the previous constraint and as such we include the average density in the weight calculation. This procedure is captured in Algorithm~\ref{alg:exploit}.

\subsection{Step 3: Enhance}

\begin{figure}[t]
    \centering
    \includegraphics[width=0.45\textwidth]{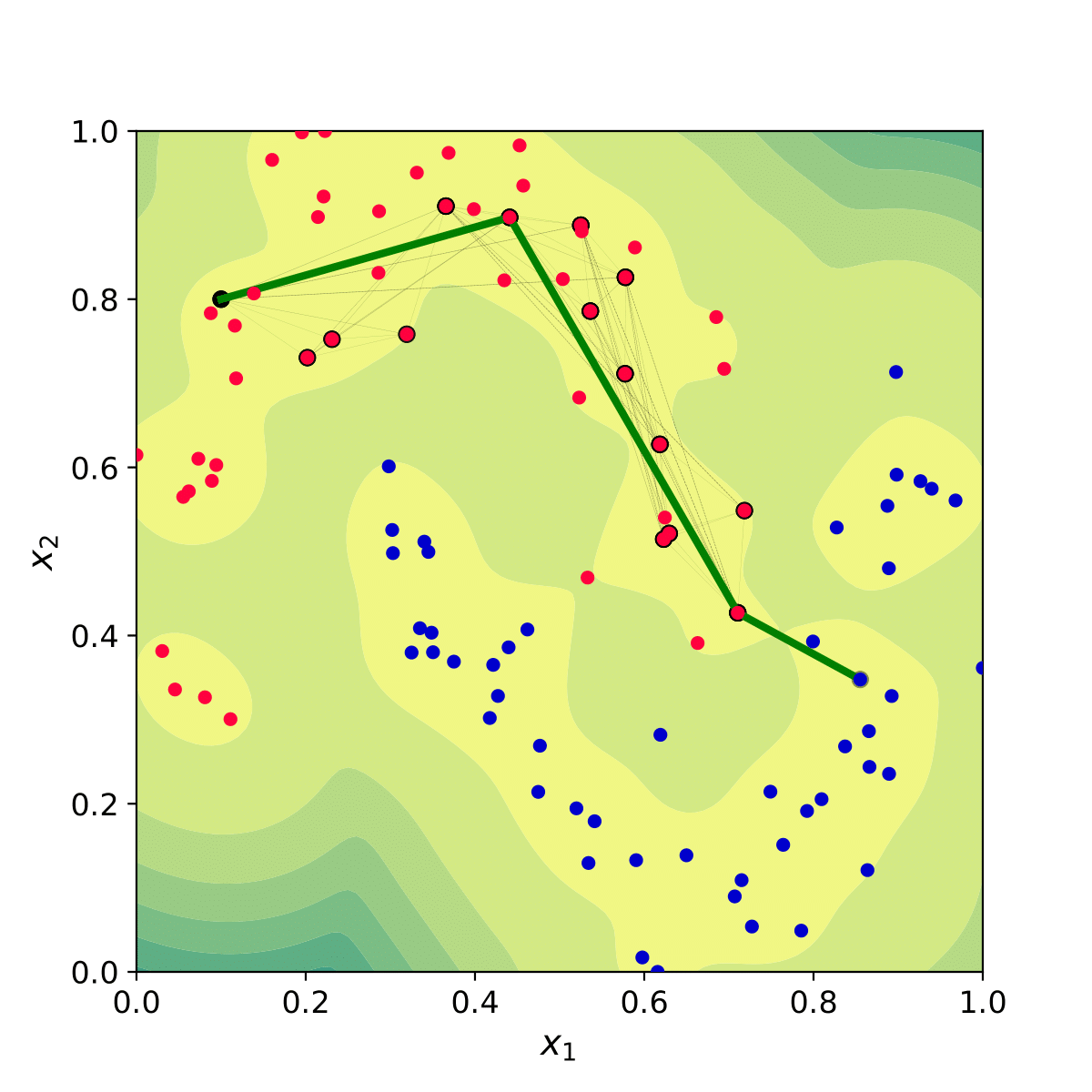}
    \caption{A local graph $G$ (black from the exploit stage) where $V\subset X$. $G$ captures the local knowledge required to find an optimal path (dark green from the enhance stage) from $x$ to $x^\prime$ through high density data regions (yellow shading).}
    \label{fig:shortest}
\end{figure}

Given a local graph $G$ such that $v_0$ is the factual and, if $N=\lvert V \rvert$ is the number of vertices, then $v_{N}$ is the counterfactual, we must find the ``shortest'' path (determined by minimum weight) between these two nodes. To this end, we can simply employ the same strategy used by FACE, i.e., the Dijkstra's shortest path algorithm. Alternatively, we could use any other shortest path strategy such as breadth first search, depth first search, A\textsuperscript{$\star$}, and the like. 
An example of this procedure is shown in Figure~\ref{fig:shortest}.

\section{Case Study: Patient Readiness for Discharge from the Intensive Care Unit}

Ensuring that patients are discharged from the intensive care unit (ICU) in a timely manner is of utmost importance for both patient safety and hospital resource allocation. Both early and delayed discharges have been shown to be detrimental to patient health~\cite{daly2001reduction, howell2011managing}. The discharge decision is complex and challenging, and, as with all decision-making across critical care, it is constrained by the cognitive burden of the clinician integrating high-dimensional physiological time-series data as well as various social and operational factors \cite{Bourdeauxe010129}. As such, numerous predictive models have been proposed to support this decision-making process~\cite{thoral2021explainable}, however for these models to be clinically useful they must be interpretable and trustworthy~\cite{sendak2020human}. Counterfactual explanations provide a way for clinicians to engage meaningfully with model predictions for decision support~\cite{10.1007/978-3-030-77211-6_38}. In this context both the privacy and the model-agnostic features of \textsc{LocalFACE} are highly desirable. 

Paired with domain expertise, \textsc{LocalFACE} could provide clinicians with feasible and actionable sequential recourse towards patient discharge from ICU. By suggesting which changes in the physiology of the deteriorated patient might enable discharge, \textsc{LocalFACE} could support clinicians in making treatment decisions. In cases of disagreement between clinician judgement and model prediction, \textsc{LocalFACE} can provide paths of counterfactual reasoning that allow the human expert to reconcile these differences in belief. In view of these synergies, our method promotes trust by providing transparency about how the clinical data set informs the model predictions.

\subsection{Method}

\begin{figure*}[t]
     \centering
     \begin{subfigure}[b]{0.32\textwidth}
         \centering
         \includegraphics[trim={10pt 10pt 10pt 10pt},clip,width=\textwidth]{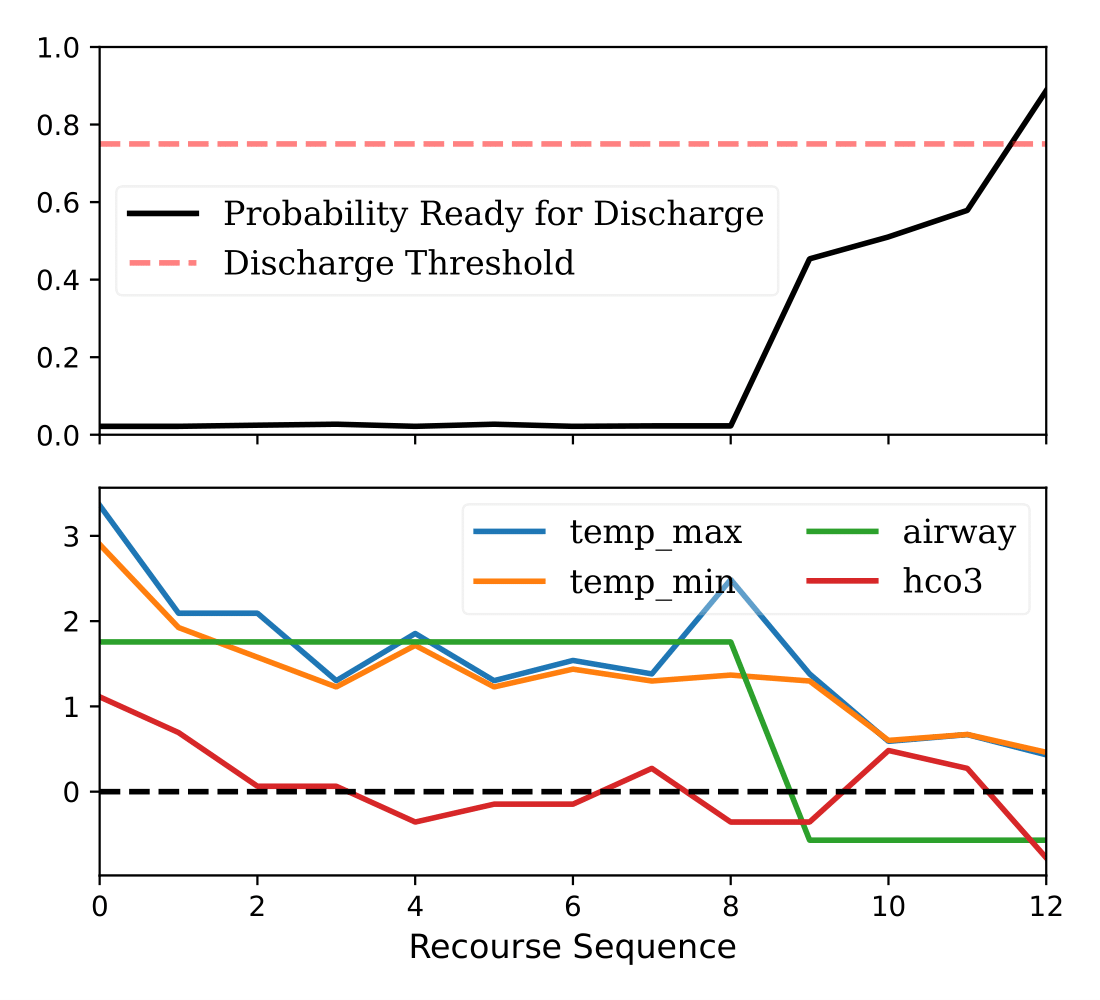}
         \caption{True negative case.\label{fig:patient_graphs:tn}}
     \end{subfigure}
     \hfill
     \begin{subfigure}[b]{0.32\textwidth}
         \centering
         \includegraphics[trim={10pt 10pt 10pt 10pt},clip,width=\textwidth]{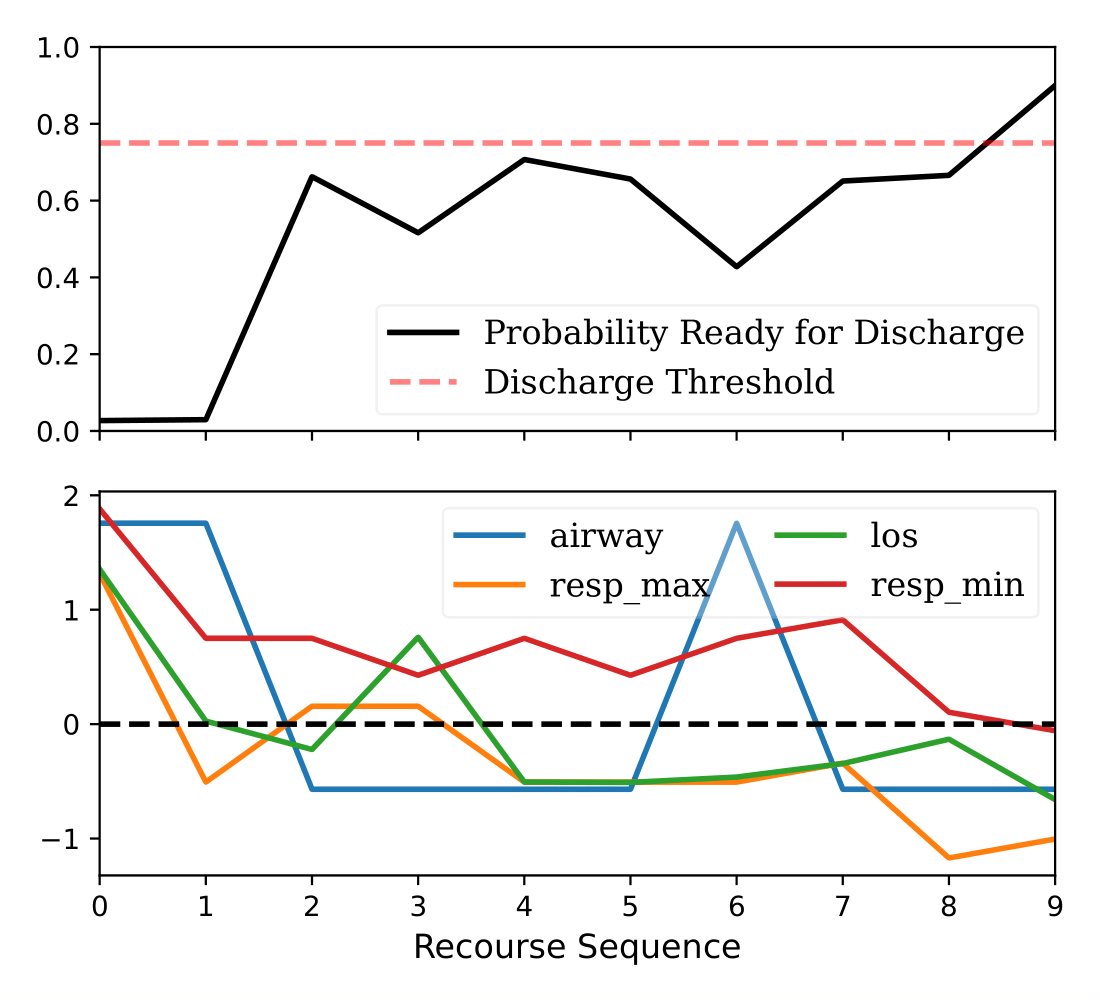}
          \caption{False negative case.\label{fig:patient_graphs:fn}}
     \end{subfigure}
     \hfill
     \begin{subfigure}[b]{0.32\textwidth}
         \centering
         \includegraphics[trim={10pt 10pt 10pt 10pt},clip,width=\textwidth]{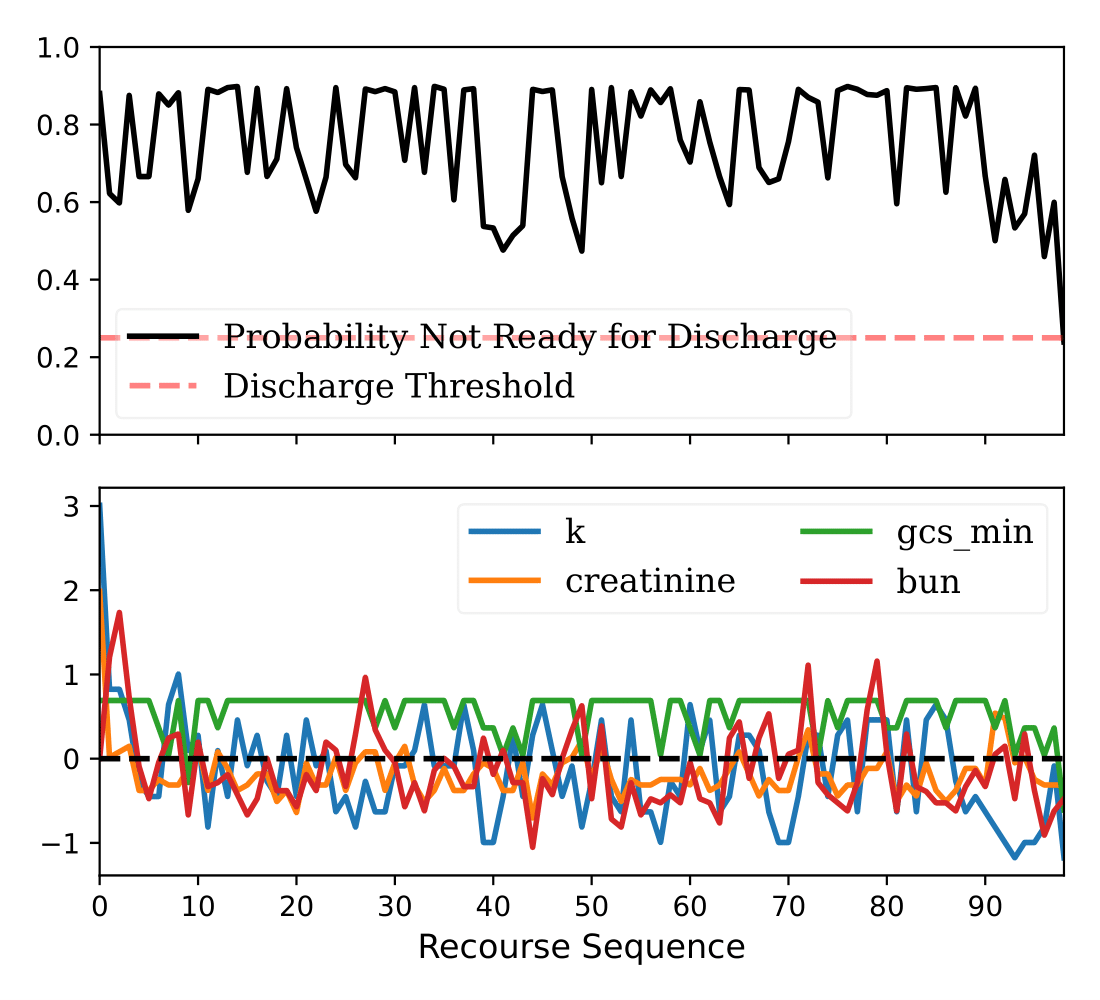}
          \caption{False positive case.\label{fig:patient_graphs:fp}}
     \end{subfigure}
     \hfill
     \caption{%
     Example patient counterfactual paths for (a) true negative, (b) false negative and (c) false positive cases from factual (recourse sequence point 0) to counterfactual (final point in recourse sequence). The readiness-for-discharge (RFD) prediction probability $[0,1]$ is given in the top panel, where a score exceeding $0.75$ implies a high certainty that a patient can be discharged. The bottom panel displays the four features that change the most from factual to counterfactual. The y-axis in the bottom panel are the scaled measurements for each feature. Tables~\ref{tab:psy_feats} and~\ref{tab:demo_feats} in Appendix~\ref{sec:extras} defines all feature abbreviations.} 
     \label{fig:patient_graphs}
\end{figure*}

We applied \textsc{LocalFACE}\footnote{\url{https://github.com/Teddyzander/localFACE}} to a pre-existing model for predicting readiness-for-discharge from intensive care~\cite{McWilliams2018TowardsAD}. The data set, which is available upon request, consists of a local cohort of 1,870 general intensive care patients (GICU) from Bristol Royal Infirmary~\cite{McWilliams2019CurationOA}, which was supplemented with a cohort of 7,592 equivalent patients from the Medical Information Mart for Intensive Care (MIMIC-III) clinical data base. Predictors, consisting of eighteen physiological features, three demographic attributes and length of stay, were sampled at the time of ``callout'', i.e., the point at which a patient was declared ready for discharge. The tables in Appendix~\ref{sec:extras} define all data features and their abbreviations. The positive class was defined as patients who left hospital alive without readmission to ICU. Such patients were deemed to be ready-for-discharge (RFD) at the time of callout, given their successful outcome. Resampling of patients from the negative class, i.e., not-ready-for-discharge (NRFD), at earlier time points during their ICU stay was used to balance the class sizes, since patients at these time points were clearly not physiologically fit for discharge. This sampling resulted in a total of 13,243 MIMIC-III and 3,506 GICU data points.

Missing feature values were imputed using a combination of forward fill and $k$-NN imputation; all features were standardised using scikit-learn's \texttt{StandardScaler}. A random forest classifier was trained to predict readiness-for-discharge on 13,943 subjects (83\% of total data), using multiple-source cross-validation to promote generalisation across the two cohorts. Grid search was used to optimise hyperparameter selection for \texttt{n\_estimators}, \texttt{max\_features}, \texttt{max\_depth}, and \texttt{band\_width}, with full details of the model being provided in the original publication. The procedure was repeated on 100 random train--test splits to produce estimates of the mean and standard deviation (SD) of classifier performance. Subjects within the test set were sampled as factuals for which \textsc{LocalFACE} was applied with $k=50$ (nearest points), allowing the algorithm to access only $0.28\%$ of the training data at each step. To aid feasibility, constraints were applied to sex, age and length of stay to ensure that only relatively comparable local points were utilised to generate the path and find the counterfactuals. Sex was considered immutable, age could vary $\pm$ c.25 years and length of stay could vary $\pm$ c.11.7 days. 

\subsection{Results}

Across 100 random train--test splits the classifier achieved mean (SD) area under the curve performance of 0.8814 (0.0060) and Brier loss of 0.1442 (0.0029). At a specificity of 0.7, the classifier achieved an F1 score of 0.8162 (0.0075), accuracy of 0.7978 (0.0067) and sensitivity of 0.8966 (0.0137). \textsc{LocalFACE} was applied to the entire test set of 2,806 patients. On average, \textsc{LocalFACE} took 0.453 seconds to complete all three stages of the algorithm on an Intel Core i7-1065G7 CPU @ 1.30GHz with 32GB of RAM, and always found a viable recourse matrix $Z$.

\subsection{Discussion}

We report a selection of individual counterfactual paths to illustrate the application of \textsc{LocalFACE} for three types of clinical use case: (1) true negative, (2) false negative, and (3) false positive. Additional examples are available in 
Appendix~\ref{sec:extras}. Where the model and clinician agree -- true negatives -- the suggested algorithmic recourse may strengthen the resolve of a clinician by backing up their claim and reasoning, or give further reasoning for why a patient is NRFD. When the model and clinician disagree -- false positives and false negatives -- the suggested algorithmic recourse may assist a clinician to more easily engage in hypothetical thought experiments to support their decision-making process, and to question their reasoning.
Allowing the clinician to more effectively query the reasoning of the model can lead to a higher level of understanding between both parties. Thus, clinicians will know when to trust the model, when to trust themselves, and when a case may be more complex than it initially appears.

\paragraph{True Negative -- Actionable Recourse to Discharge}
In this use case, the patient is NRFD and correctly classified as such by the random forest classifier. The question of clinical interest is how to bring the patient into a physiological state in which they are ready to be discharged from critical care.
Figure~\ref{fig:patient_graphs:tn} shows such a patient and the recourse sequence, which was constructed using 1.69\% of the training data, required to make this patient RFD. The patient begins a long way from RFD, as indicated by the low RFD probability and elevated temperature readings, the latter of which suggests a fever. By bringing down the temperature (``temp\_max'' and ``temp\_min'') and serum bicarbonate (``hco3''), and weaning the patient off mechanical ventilation (``airway''), the patient can be classified as RFD. This path is both actionable and clinically meaningful~\cite{knaus1985apache, paudel2022serum}.

In general, we find that the observed counterfactual paths are easily interpreted in terms of the original patient physiology. For example, actions such as weaning from mechanical ventilation and improvement of physiological measurements are both strong requirements for de-escalation of critical care~\cite{kajdacsy2005use}, and are identified as such by \textsc{LocalFACE} for relevant patients.

\paragraph{False Negative -- Hypothetical Recourse to Discharge}

In this case we present a patient who was classified as NRFD, but who was deemed RFD by a consultant and discharged with a successful outcome. Such cases represent 3.8\% of the test set (107 total patients). Here the clinician might ask: ``What recourse according to the model is required to bring this patient to RFD?''.
Figure~\ref{fig:patient_graphs:fn} shows a patient whose NRFD classification and recourse sequence are dominated by respiratory concerns.  The recourse was constructed using just 1.27\% of the training data. Utilising the classifier, \textsc{LocalFACE} suggests that weaning off mechanical ventilation (``airway''), associated reduction in respiratory rate (``resp\_max'' and ``resp\_min'') and reduction in length of stay (``los'') would enable ICU discharge. Although a reduction in length of stay is not actionable, its inclusion in recourse may provide useful insights for clinical reflection, for example: ``If this was on day 3 of their stay, we would be saying they were ready for discharge, so maybe we should think about doing so.'' The recourse sequence briefly includes reintubation at step 6 (``airway''), corresponding to an appropriate local drop in RFD score. Reintubation following weaning from mechanical ventilation is not uncommon clinically~\cite{listello1994unplanned}.

From the available features it is difficult to see how this patient was considered RFD by the clinician as the patient is in respiratory failure~\cite{fujishima2014pathophysiology}. The counterfactual reasoning provided by \textsc{LocalFACE} may prompt them to reconsider their decision. In other scenarios (Figure~\ref{fig:supp_fn_7}, Appendix) one clear clinically meaningful step is required for counterfactual RFD classification. If the clinician knows that these vital signs are improving with a convincing trend, \textsc{LocalFACE} may provide reassurance that RFD is imminent and that the clinician can be confident in sticking with their assessment of RFD.

\paragraph{False Positive -- Reconciling Differing Beliefs}
This case is similar to the false negative case above in that there is a disagreement between the classifier output (RFD) and the clinician's decision (NRFD). The clinician may want to confirm if the counterfactual reasoning is strong enough for them to change their decision and accept the RFD recommendation of the classifier. To do this we reverse the RFD classification of the factual to result in recourse towards an NRFD outcome. False positives represent 15.4\% of the test set (433 total patients).
Figure~\ref{fig:patient_graphs:fp} shows such a patient sequence with recourse calculated using 5.18\% of the training data. It is not until consciousness (``gcs\_min'') and potassium level (``k'') are reduced that the patient is classified as NRFD. The required length of the sequence suggests that the patient exists in a dense region of other RFD data points far from the decision boundary. It is likely that other, non-physiological considerations available to the clinician but not the classifier were driving their decision to keep the patient in ICU~\cite{stelfox2015scoping}. \textsc{LocalFACE} may prompt the clinician to consider if these outside factors are strong enough to warrant not discharging the patient when they are physiologically RFD.

\section{Related Work}

Counterfactual explainability is one of the most popular XAI techniques given its strong human-centred foundations~\cite{miller2019explanation}. %
Such explanations also comply with various legal frameworks, which further adds to their appeal~\cite{WachterSandra2017CEWO}. %
Within this framework, algorithmic recourse~\cite{ustun2019actionable} attempts to align counterfactuals with the iterative explainability paradigm, which is second nature to humans. 
While many technical realisations of this concept exist~\cite{10.1145/3442188.3445899}, they exhibit various shortcomings as we have shown through this paper. %

Similar in spirit to our approach yet distinct, some explanation tools leverage a corpus of examples from the training data to explain predictions of a model~\cite{NEURIPS2021_65658fde}. Here, algorithmic recourse can be thought of as a collection of exemplars that transform an individual assigned an undesired prediction into an instance classified as desired. Referring back to our case study, a clinician can use such an explanation to determine what features are the most important to change, how they should change and, crucially, in what sequence they should change in order to maximise the likelihood of a successful outcome.
However, in contrast to our approach, the burden of distilling this information is with the clinician who may lack the technical (AI and ML) expertise required to this end~\cite{sokol2023reasonable}.

\section{Conclusion}

This paper introduces \textit{sequential algorithmic recourse}, which constructs a piece-wise path from a factual to a counterfactual instance. We achieve this by considering local steps guided both by the density function, which is a proxy for feasibility, and an increase in the predicted posterior class probability, which is a proxy for improving odds of positive classification. Considering these aspects allows the recourse to capture more complex dynamics such as when certain actions should be taken, in what order they should occur, and what actions cannot be completed in parallel. We argue that such a perspective on algorithmic recourse -- the consideration of a set of multiple actions in order to achieve a desired outcome -- is well aligned with human thinking. Thus, the counterfactual retrieved by our method is more easily scrutinised by experts, as is the decision-making process of the model. This is also well aligned with other works such as feature importance and exemplars, where sequential algorithmic recourse considers examples along the proposed path in order to explain suggested changes. Notably, sequential algorithmic recourse allows an expert to effectively and intuitively understand if a given counterfactual is a good example for the queried factual, which we demonstrated for discharging ICU patients.

Future work will include known traversed paths into the training set in order to compare the proposed recourse to previously completed recourse. Additionally, we will critically assess the usefulness of algorithmic recourse in scrutinising automated decisions with ICU clinicians. It would be especially interesting to investigate cases where a model disagrees with a clinician, which may prompt attempts to reconcile why this is the case. Knowing which decision to trust and distrust will assist in shouldering the decision-making burden, and will allow clinicians to better distribute mental load, saving energy for more complex tasks. Since vital signs can be affected by organ support, medication and other interventions, which information is not available to us at present, we will look into incorporating it as it could strengthen both the model and explanatory recourse.

\section{Acknowledgments}
Edward A. Small, Jeffrey Chan and Flora D. Salim were supported by the ARC Centre of Excellence for Automated Decision-Making and Society (project number CE200100005), funded by the Australian Government through the Australian Research Council. KS was supported by the Hasler Foundation, grant number 21050. Jeffrey N. Clark and Raul Santos-Rodriguez are funded by the UKRI Turing AI Fellowship EP/V024817/1. Chris McWilliams was funded by NIHR through the NHS AI Lab’s AI in Health and Care Award (AI\_AWARD01943). The views expressed are those of the author(s) and not necessarily those of the NIHR, the Department of Health and Social Care or NHS England.

\bibliography{aaai22}

\cleardoublepage\newpage

\renewcommand\thefigure{\thesubsection.\arabic{figure}}

\appendix
\section{Additional Case Study Examples}
\label{sec:extras}

\begin{table*}[t]
\centering
\begin{tabular}{llll}
\hline
Test & Variable                                            & Abbreviation & Test condition                                                  \\ \hline
1    & Respiratory: airway                                 & airway       & airway patent                                                   \\
2    & Respiratory: Fio2                                   & fio2         & fio2$\leq$0.6                                                        \\
3    & Respiratory: blood oxygen                           & spo2\_min    & spo2$\geq$95 (\%)                                                    \\
4    & Respiratory: bicarbonate                            & hco2         & hco3$\geq$19 (mmol/L)                                                \\
5    & Respiratory: rate, minimum                          & resp\_min    & 10$\leq$resp (bpm)                                                   \\
6    & Respiratory: rate, maximum                          & resp\_max    & resp$\leq$30 (bpm)                                                   \\
7    & Cardiovascular: systolic blood pressure, minimum    & bp\_min      & bp$\geq$100 (mm Hg)                                                  \\
8    & Cardiovascular: heart rate, minimum                 & hr\_min      & 60$\leq$hr (bpm)                                                     \\
9    & Cardiovascular: heart rate, maximum                 & hr\_max      & hr$\leq$100 (bpm)                                                    \\
10   & Central nervous system: Pain                        & pain         & 0$\leq$pain$\leq$1                                                        \\
11   & Central nervous system: Glasgow coma score, minimum & gcs\_min     & gcs$\geq$14                                                          \\
12   & Central nervous system: Temperature, minimum        & temp\_min    & 36$\geq$temp (°C)                                                    \\
13   & Central nervous system: Temperature, maximum        & temp\_max    & temp$\leq$37.5 (°C)                                                  \\
14   & Bloods: haemoglobin                                 & haemoglobin  & haemoglobin$\geq$90 (g/L)                                            \\
15   & Bloods: potassium                                   & k            & 3.5$\leq$k$\leq$6.0 (mmol/L)                                              \\
16   & Bloods: sodium                                      & na           & 130$\leq$na$\leq$150 (mmol/L)                                             \\
17   & Bloods: creatinine                                  & creatinine   & 59$\leq$creatinine$\leq$104 (umol/L)                                      \\ 
18   & Bloods: urea                                        & bun          & \begin{tabular}[c]{@{}l@{}}2.5$\leq$bun$\leq$7.8 (mmol/L)\end{tabular} \\ \bottomrule
\end{tabular}
\caption{Codified version of the discharge criteria for application to electronic health record data. Here the 15 criteria have
been grouped into intuitive subsets, with abbreviations attribute to each variable. According to the original specification, if all 18
criteria are met for a period of at least 4 hours the patient can be safely discharged. Table taken with author's consent~\cite{McWilliams2018TowardsAD}.}
\label{tab:psy_feats}
\end{table*}

\begin{table*}[t]
\centering
\begin{tabular}{@{}ll@{}}
\toprule
Variable                               & Abbreviation \\ \midrule
Demographics: age at admission         & age          \\
Demographics: sex                      & sex          \\
Demographics: body mass index          & bmi          \\
Hours since admission (length of stay) & los          \\ \bottomrule
\end{tabular}
\caption{Patient characteristics for the two cohorts.}
\label{tab:demo_feats}
\end{table*}

\subsection{True Negative}
\setcounter{figure}{0}
\label{sex::tn}
\begin{figure}[!ht]
    \centering    \includegraphics[width=0.40\textwidth]{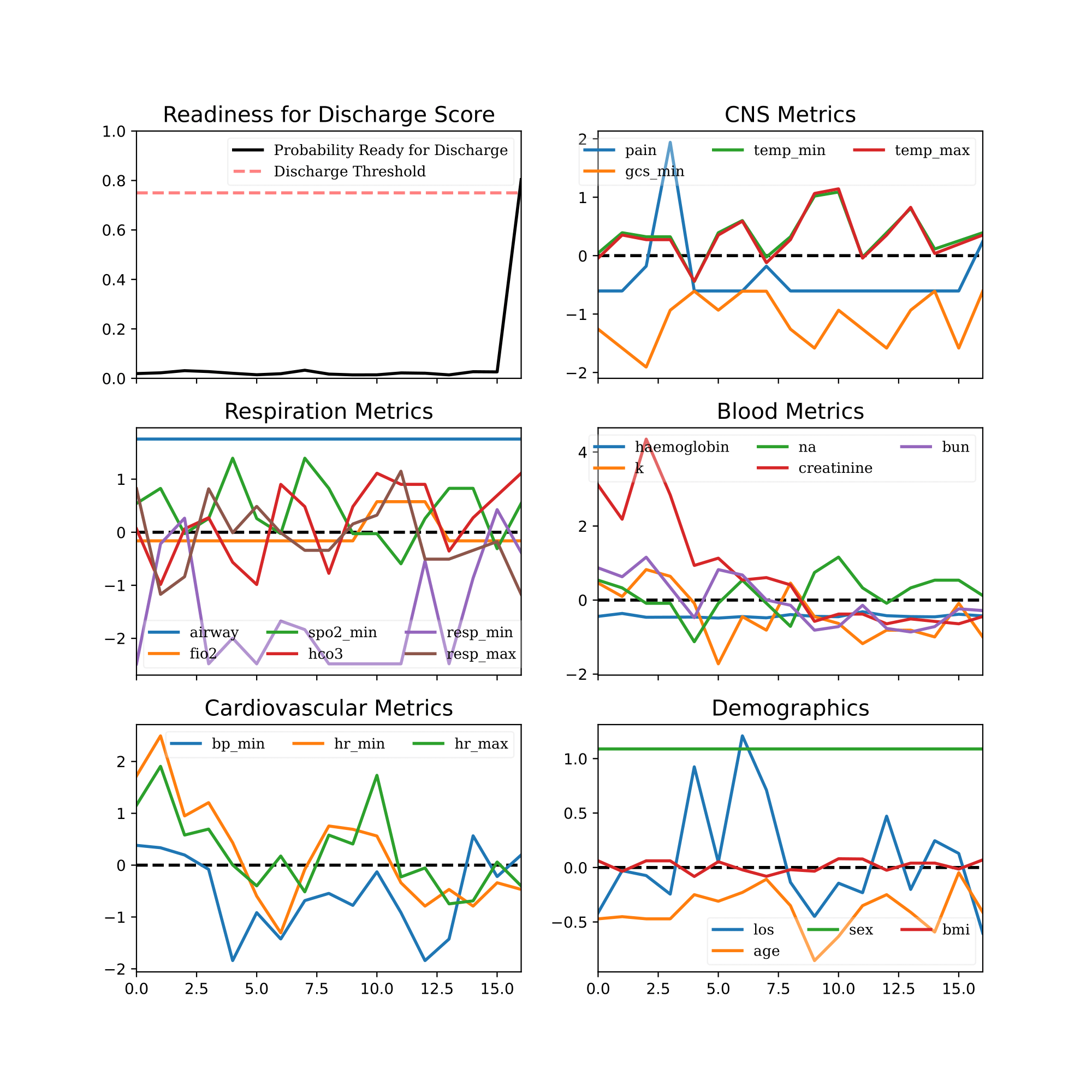}
    \caption{True negative example patient factual from NRFD classification to RFD classification.
    The patient is a long way from being RFD. They require weaning from ventilation (airway). Creatinine, serum sodium (na) and heart rate need to reduce. The patient may have kidney disease.
    }
    \label{fig:supp_tn_1}
\end{figure}

\subsection{False Negative}
\setcounter{figure}{0}
\label{sex::fn}

\begin{figure}[!h]
    \centering    \includegraphics[width=0.40\textwidth]{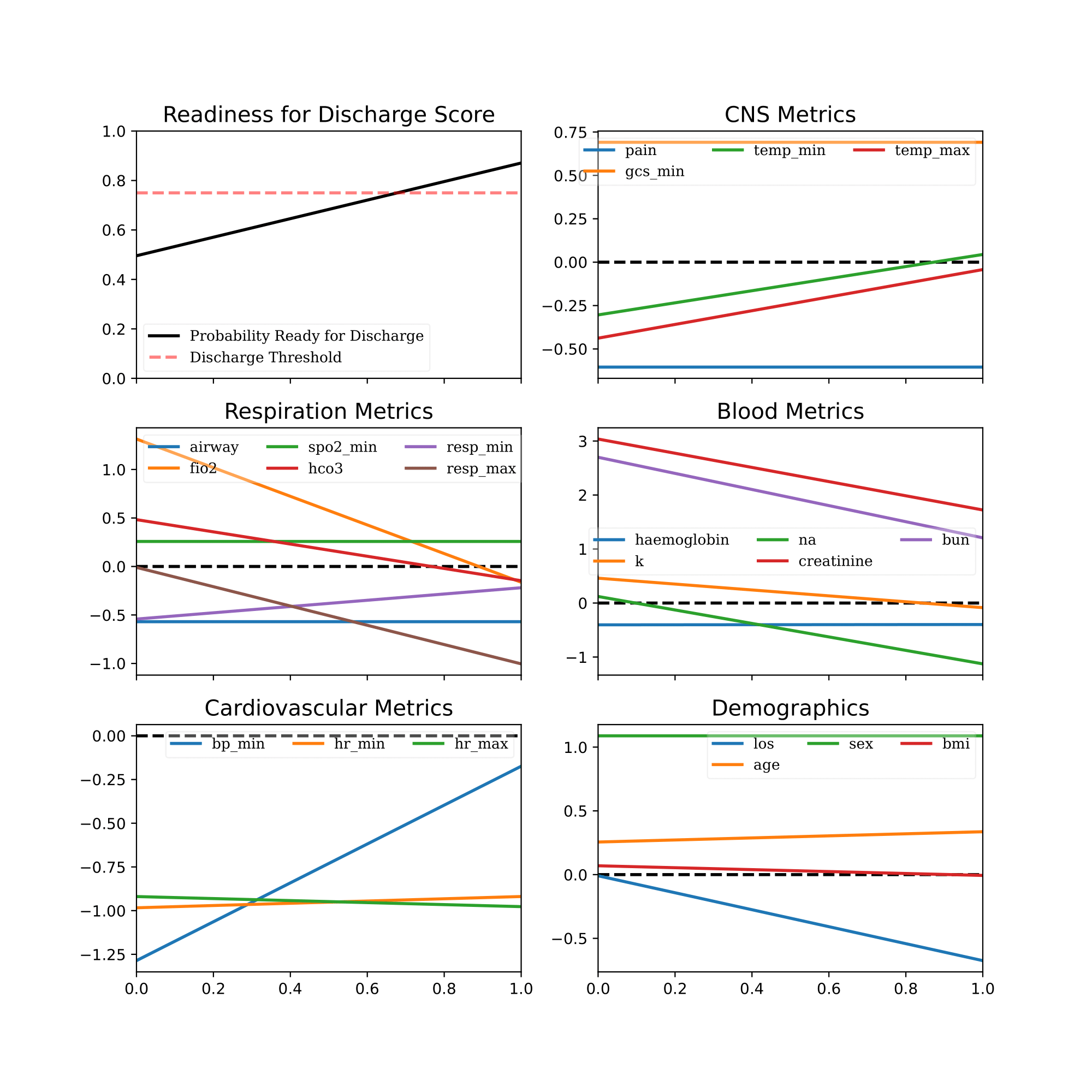}
    \caption{False negative example patient factual requiring one clear step to achieve RFD. Kidneys and hydration need to improve and oxygen dependence to come down. In false negative cases such as this the clinician would likely be confident to stick with their assessment of RFD for such a patient, especially if the trend of physiological measurements was going in the right direction. 
    }
    \label{fig:supp_fn_7}
\end{figure}

\subsection{False Positive}
\setcounter{figure}{0}
\label{sex::fp}
\begin{figure}[!h]
    \centering    \includegraphics[width=0.40\textwidth]{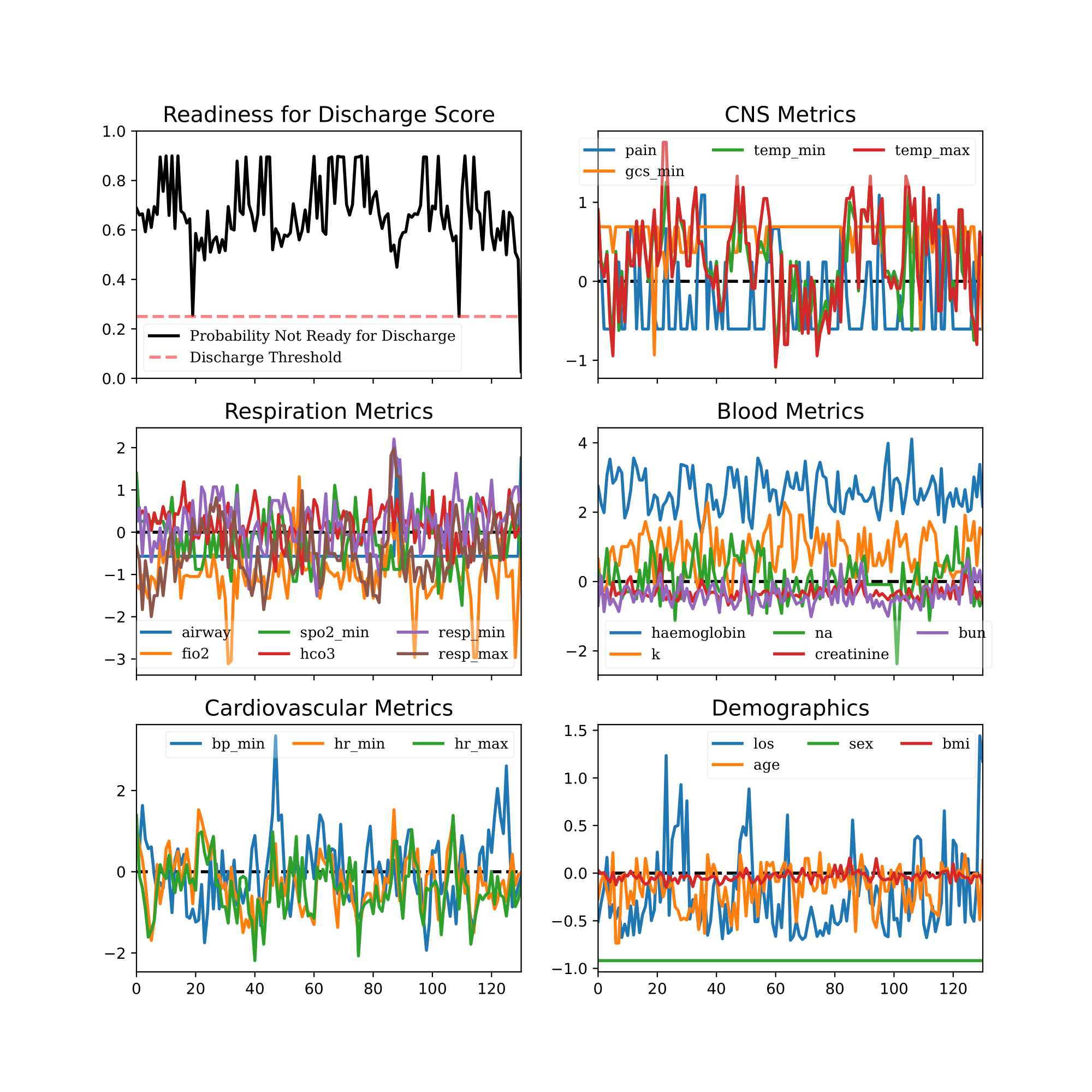}
    \caption{False positive example patient factual path from RFD classification to NRFD classification. From the presented data it seems reasonable that the patient was classified RFD. The alteration to result in NRFD makes sense: worse gcs, fio2, and cardiovascular metrics. An unexpected event/deterioration (leading to readmission or mortality) following discharge appears likely. If the outcome were not known, it is likely that clinician would stick with their decision of NRFD. 
    }
    \label{fig:supp_fp_1}
\end{figure}

\section{Minimum Angle Proof}
\setcounter{figure}{0}
\label{sec:min_angle}

During the exploration stage of \textsc{LocalFACE} we can allow the search to drift away from specific data points provided that we remain within a tolerance $\epsilon$. This allows for a degree of flexibility in the search phase whilst still remaining within the areas where data exist. It is therefore crucial that we specify criteria to construct the accessible space during each step of the search process. To this end, we utilise some fundamental aspects of geometry and linear algebra.

Assume there exists a distance function $d: \mathcal{X} \times \mathcal{X} \mapsto \mathbb{R}$. We wish for the line $L$ that connects any point $x$ to the point $x_t$ to never be more than $\epsilon$ away from any point in $X$. In other words, $d(x_t, L(x)) < \epsilon$ for all $x \in X$. We therefore search for a relationship between the line $L$ that connects $x_t$ to $x$ and the line $L_X$ that connects the point $x$ to the nearest data point in $X$ to $x_t$ in order to construct the accessible area as shown in Figure~\ref{fig:dist} and outlined by Theorem~\ref{pythagorean}.

\begin{theorem}[Maximum Angle of Deviation]
\label{pythagorean}
Consider two hyper-spheres of radius $\epsilon$ each centred on the points $x_1,x_2\in\mathcal{X}$ such that, if $d:\mathcal{X}\times\mathcal{X}\mapsto\mathbb{R}$ is a distance function, then $d(x_1,x_2) \leq 2\epsilon$. Consider $L_x$ to be the line that connects $x_1$ and $x_2$, and $L$ to be any line that connects $x_1$ to $x_t$ such that $d(x_2,x_t) \leq \epsilon$. In order for $d(x, x_1) \leq \epsilon$ or $d(x, x_2) \leq \epsilon$ for all $x$ on $L$ then
\begin{equation*}
    \phi \geq 
    \frac{1}{2}\Big(1 + \frac{d}{\epsilon}\Big)
\end{equation*}
where
\begin{equation*}
    \phi = \frac{(x_2 - x_1) \cdot (x_t - x_1)}{\lVert x_2 - x_1 \rVert \lVert x_t - x_1 \rVert}
    \text{~.}
\end{equation*}
\end{theorem}

\begin{figure}[t]
    \centering    \includegraphics[width=0.4\textwidth]{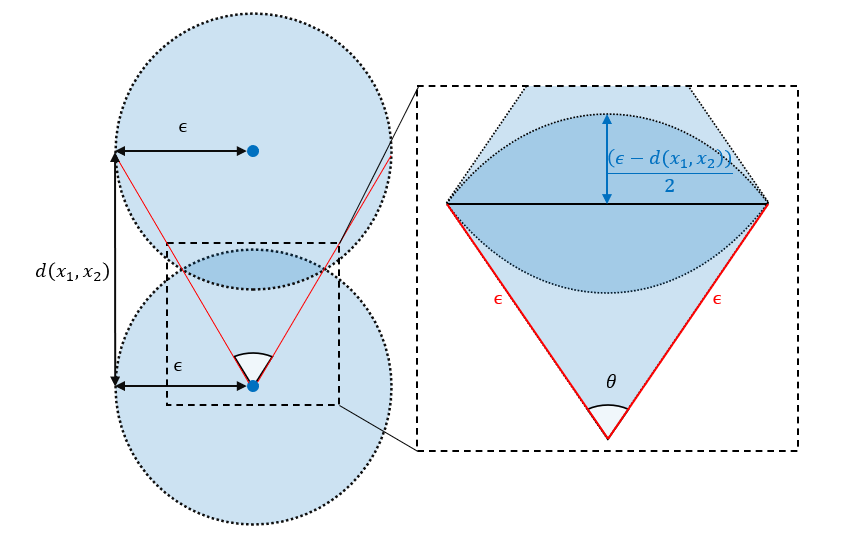}
    \caption{The maximum angle of a trajectory $\theta$ can be calculated using known values of the distance $d$ and tolerance $\epsilon$. We must then ensure that the angle between $L$ and $L_x$ is within $\frac{1}{2}\theta$.}
    \label{fig:angle}
\end{figure}

\begin{proof}
Consider the set up provided in the theorem. We have two points $x_1$ and $x_2$ such that the distance between them is $d$. Centred on each point is a hyper-sphere of radius $\epsilon$ such that $d < 2\epsilon$. Thus, the hyper-spheres must overlap. The plane that passes through each of the points where the hyper-sphere surfaces intersect creates a chord, with the height of the arced portion for each hyper-sphere given by
\begin{equation}
    h=\frac{1}{2}(\epsilon - d)
    \text{~.}
    \label{eq:h}
\end{equation}
The two planes that intersect at the points where the hyper-spheres touch and $x_1$ create an angle $\theta$. Therefore, if $v=x_2-x_1$ and $u=x_t-x_1$, then we must ensure that the angle between $v$ and $u$ is less than $\frac{1}{2}\theta$, or
\begin{equation}
    \frac{v \cdot u}{\lVert v \rVert \lVert u \rVert} \geq \cos\Big({\frac{1}{2}\theta}\Big)
    \text{~.}
    \label{eq:angle_cond}
\end{equation}
From the definition of segment height (sagitta)~\cite{10.1007/978-3-642-13728-0_7}:
\begin{equation*}
\begin{aligned}
        h &= \epsilon\bigg(1-\cos{\Big(\frac{\theta}{2}\Big)}\bigg) \\
        \implies \theta &= 2\arccos\bigg({1 + \frac{h}{\epsilon}}\bigg)
        \text{~.}
\end{aligned}
\end{equation*}
Substituting this into Equation~\ref{eq:angle_cond} gives
\begin{equation*}
    \begin{aligned}
        \frac{v \cdot u}{\lVert v \rVert \lVert u \rVert} &\geq \cos\Big({\frac{1}{2}2\arccos\bigg({1 - \frac{h}{\epsilon}}\bigg)}\Big) \\
        \frac{v \cdot u}{\lVert v \rVert \lVert u \rVert} &\geq 1 - \frac{h}{\epsilon} 
        \text{~.}
    \end{aligned}
\end{equation*}
Then, from Equation~\ref{eq:h}
\begin{equation*}
    \begin{aligned}
        \frac{v \cdot u}{\lVert v \rVert \lVert u \rVert} &\geq 1 - \frac{\frac{1}{2}(\epsilon - d)}{\epsilon} \\
        &\geq 1 - \frac{1}{2}\bigg(\frac{\epsilon}{\epsilon} - \frac{d}{\epsilon}\bigg) \\
        &\geq 1 - \frac{1}{2} + \frac{d}{2\epsilon} \\
        &\geq \frac{1}{2}\bigg(1+\frac{d}{\epsilon}\bigg)
    \end{aligned}
\end{equation*}
and therefore
\begin{equation}
    \frac{(x_2 - x_1) \cdot (x_t - x_1)}{\lVert x_2 - x_1 \rVert \lVert x_t - x_1 \rVert} \geq \frac{1}{2}\bigg(1+\frac{d}{\epsilon}\bigg)
        \text{~.}
\end{equation}    
\end{proof}

Thus, when $d=\epsilon$, $L$ and $L_x$ must be perfectly aligned (normalised dot product of $1$). When
\begin{equation*}
    \frac{d}{\epsilon} \leq \pi - 1
        \text{~,}
\end{equation*}
we can simply check that $d(x_t,x_2) \leq \epsilon$ as this implies $d(x, x_t) \leq \epsilon$ for all $x$ on $L$. A visual interpretation of this relation is shown in Figure~\ref{fig:angle}.

\end{document}